# EFFECTIVE USER INTENT MINING WITH UNSUPERVISED WORD REPRESENTATION MODELS AND TOPIC MODELLING

Bencheng Wei[1]

[1]Smith School of Business, Queen's University, Kingston, Ontario
20bw3@queensu.ca

*ABSTRACT*

*Understanding the intent behind email/chat between customers and customer service agents has become a crucial problem nowadays due to an exponential increase in the use of the Internet by people from different cultures and educational backgrounds. More importantly, the explosion of e-commerce has led to a significant increase in text conversation between customers and agents. In this paper, we propose an approach to data mining the conversation intents behind the textual data. Using the customer service dataset, we train unsupervised text representation models using continuous bag of words (CBOW) and Skip-Ngram, and then develop an intent mapping model which would rank the pre-defined intents base on cosine similarity between sentences' embeddings and intents' embeddings. Topic-modeling techniques are used to define intents and domain experts are also involved to interpret topic modelling results. With this approach, we can get a good understanding of the user intentions behind the unlabelled customer service textual data.*

## 1. INTRODUCTION

Great amount of customer interactions such as call summaries, email requests, and meeting notes are generated daily by customer service agents. These catalogued interactions are a treasure trove of information. They contain success stories, pain points, and even suggestions from customers to help improve the products or services. However, for these conversation data, it is not easy to understand the main intent behind the text. It is a manually intensive process that requires a lot of time and dedicated resources to label intents for these conversations. The vast majority of companies with Customer Relationship Management (CRM) tools have their agents manually tag email interactions today. This manual process turns unstructured text into structured data and enables management and other stakeholders to identify trends and opportunities behind textual data.

Therefore, we want to leverage NLP and Deep Learning to build unsupervised models to analyze intents behind textual conversations. This model will empower businesses to aggregate, analyze and gain valuable insights from these documented transactions, and allow management or internal partners to gather information about customer personas and product feedback, as well as increase the throughput of the service agents. In this paper, we are going to demonstrate how we build unsupervised intent mining models to define intents and map these predefined intents to textual conversations. Topic modeling is used to understand intents and word representation technique Word2Vec is used to turn text data into vector representations.

## 2. DATA SUMMARY

In this paper, we use private data provided by Artwo AI. The data is stored as a Comma-separated values (CSV) file in Amazon AWS S3 and there are 3 days of customer service support tickets. There are around 110k support tickets per day. Each ticket is concatenated with multiple emails, chats between customers and agents. Because a ticket will not be closed until the issue is resolved. As a result, a ticket might contain multiple intents. Moreover, metadata of these tickets is also provided which contain information of ticket subject, open dates, priority, etc.

The biggest challenge we face is the quality of data. The dataset is enormous and contains three types of data including Calls, chats, and emails. There are numerous patterns, grammatical errors, and spellings errors in the corpus. For example, each email carries a disclaimer and each agent uses a slight variation of the same disclaimer. Moreover, there are many systems generated text that needs to be handled and removed. Issues like these have caused challenges in cleaning data.

## 3. TEXT PREPROCESSING

The preprocessing of the text data is an essential step as it makes the raw text ready for mining. If we skip this step, there is a higher chance that you are working with noisy and inconsistent data. Text data coming from different sources usually have different characteristics, which makes text preprocessing as one of the most important steps in the nature language processing (NLP) pipeline. For instance, text data from Twitter is significantly different from text data on Stackoverflow, or news blogs. Thus, different sources of text are required to be treated differently.

### 3.1. Content extraction and Pattern removal

The main objective of this step is to extract the main content from the chat ticket and remove repeat system generated patterns, email disclaimers and attachments in the emails. Term frequency mining is used to explore the patterns that appear frequently in most of the tickets. For example, each email carries a disclaimer and each agent uses a slight variation of the same disclaimer. What's more, there are system generated texts in different forms of ticket. These systems generate texts are meaningless to be included into modelling.

### 3.2. Text cleansing

To further clean the text, we took the following steps to overcome the challenges and characteristics of the data described above. Below are a couple examples we address during preprocessing:

- Conversion of text to all lower-case characters
- Removal of special characters and super long characters
- Conversion of different or non-standard language into English
- Elimination of markups and Removal of URLs using regex
- Elimination of email attachments and signatures
- Conversion of non-textual elements like emojis into text
- Stopword removal (Use for Topic Modelling and Intent Mapping)
- Lemmatize

We did not use some of the traditional data cleaning techniques such as stemming and Stopwords removal when we train the word representation, as it might change the meaning of the word by removing parts of the words, resulting in inaccurate intent analysis. Therefore, we ignored such techniques and only used these additional techniques when we run topic modelling and intent mapping.

# 4. WORD REPRESENTATION

## 4.1. Word2Vec with CBOW and Skip-Ngram

Word2Vec is a widely used word representation technique which leverages neural networks under the hood. The resulting word representation can be used to analyze semantic similarity between words and phrases.

Continuous Bag-of-Words Model (CBOW) which predicts the middle word based on surrounding context words. The context consists of a few words before and after the current (middle) word. This architecture is called a bag-of-words model as the order of words in the context is not important. Continuous Skip-Ngram Model (Skip-Ngram) which predicts words within a certain range before and after the current word in the same sentence.

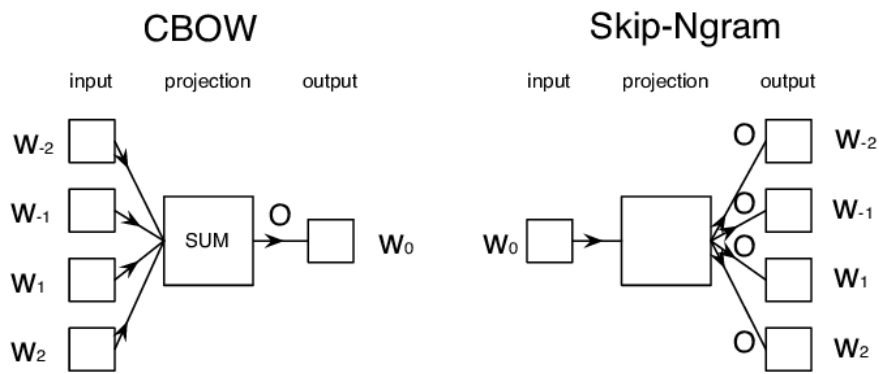

Figure 1. CBOW and Skip-Ngram Architecture

If we look at the Skip-Ngram model, it is a model that uses a neural network of one hidden layer to predict context words w(2), w(1), w(-1), w(-2) of an input word w(0). In the other words, the model attempts to maximize the probability of observing all four surrounding words together, given a center word. The training objective is to learn word vector representations that are good at predicting nearby words.

Mathematically, it maximizes the probability of predicting context words leads to optimizing the weight matrix that best represents words in a vector space.

Due to the difference of neural network architecture design, CBOW learns better syntactic relationships between words while Skip-Ngram is better in capturing semantic relationships. In practice, for the word 'dogs' CBOW would retrieve as closest vectors morphologically similar words like plurals, i.e. 'dogs' while Skip-gram would consider morphologically different words (but semantically relevant) like 'cat'.

When we train a Word2Vec model, we won't be interested in the inputs and outputs of the neural network, rather the goal is actually just to learn the weights of the hidden layer that are actually the "word vectors" that we're trying to learn.

The model we present in this paper is based on Skip-Ngram since we want our model to detect semantic relations between words, which we believe is more appropriate for intent mining use cases. Below is the T-Sne visualization of the word representation. We can see there are clusters of related words in the figure below. For example, item quality related words including "Leakage", "Wet", "Damaged", "Crushed", "Unwrapped", "Opened", "Cracked" are close to each other in 2-dimension word representation.

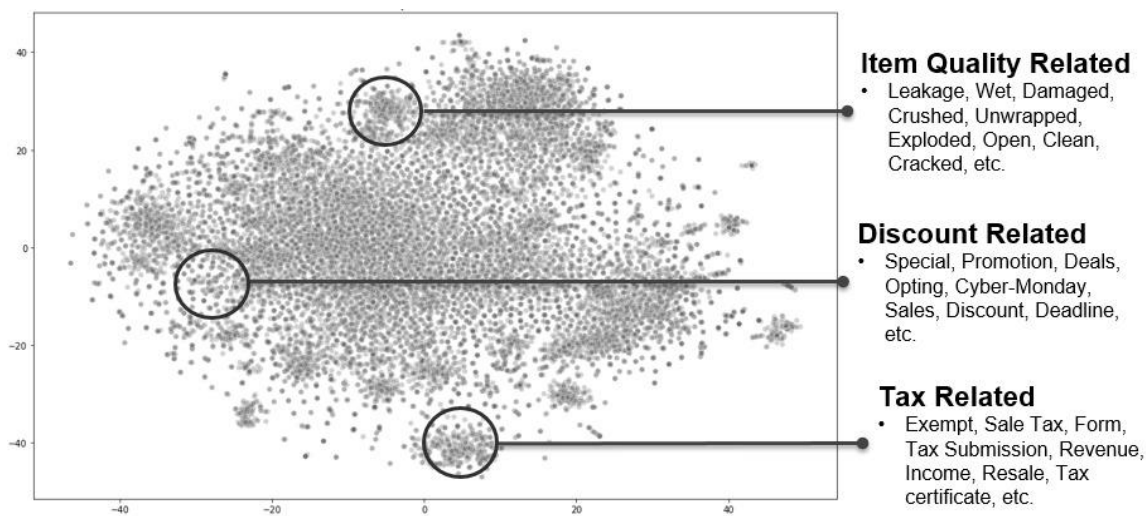

Figure 3. Spam traffic sample

### 4.2. Measure Similarity Between Texts

Cosine Similarity is one of the popular metrics to measure the text-similarity between two documents irrespective of their size. A word can be represented into a vector form and text documents are represented in n-dimensional vector space. Mathematically, the cosine similarity metric measures the cosine of the angle between two n-dimensional vectors projected in a multi-dimensional space. The mathematical equation of Cosine similarity between two non-zero vectors is:

$$\text{similarity} = \cos(\theta) = \frac{\mathbf{A} \cdot \mathbf{B}}{\|\mathbf{A}\|\|\mathbf{B}\|} = \frac{\sum_{i=1}^{n} A_i B_i}{\sqrt{\sum_{i=1}^{n} A_i^2} \sqrt{\sum_{i=1}^{n} B_i^2}},$$

Equation 1. Cosine Similarity formula

In order to measure word similarity, we calculate the word embeddings and directly measure two word's similarity. Below is an example of showing the neighbors word of 'Late' based on the Skip-Ngram model and CBOW model. Both CBOW and Skip-Ngram models are trained on 110k tickets.

Table 1. Neighbours Words of Late

| Neighbours Words | Cosine Similarity (Skip-Ngram) | Neighbours Words | Cosine Similarity (CBOW) |
|---|---|---|---|
| later | 0.748 | later | 0.753 |
| wait | 0.741 | laterly | 0.736 |
| postponed | 0.719 | sate | 0.687 |
| overloaded | 0.683 | pliate | 0.649 |
| annoyed | 0.661 | sooner | 0.632 |
| early | 0.637 | today | 0.628 |
| apologies | 0.622 | postponed | 0.615 |

For measuring sentence similarity, we calculate the sentence embeddings and then average these sentence embeddings before we measure two vector's cosine similarity. When we average the embeddings for a sentence, sentence embeddings will divide each word vector by its norm and then average them.

## 5. DEFINE INTENTS USE CASE

Since we don't have labels to train the supervised model, we are using topic modelling techniques and ticket metadata to analyze topics of intent domains. We have combined both domain knowledge, metadata and topic modelling to define the related intent topics in our case.

### 5.1. Topic Modelling

In machine learning and natural language processing, topic models are generative models, which provide a probabilistic framework. Topic modelling methods are generally used for automatically organizing, understanding, searching, and summarizing large electronic archives. The "topics" signifies hidden, to be estimated, variable relations that link words in a vocabulary and their occurrence in documents. A document is seen as a mixture of topics. Topic models discover the hidden themes throughout the collection and annotate documents according to those themes. Each word is seen as drawn from one of those topics. Finally, a document coverage distribution of topics is generated and it provides a new way to explore the data on the perspective of topics.

In order to generate topics from the documents, there are a couple ways including Latent Semantic Analysis (LSA), Probabilistic Latent Semantic Analysis (PLSA), Latent Dirichlet Allocation (LDA), Non Negative Matrix Factorization (NMF), etc. Recently, there are also new models using pre-trained embeddings to get topics from documents. BERTopic is an example that leverages transformers and creates dense clusters allowing for easily interpretable topics.In this paper, we use LDA as our main topic modelling method. There is no best practice to conclude which method is the best method to do topic modelling. Because topic modelling is an unsupervised technique and there is no accuracy we can compare across different methods.

When running topic modelling, one of the problems we need to answer is how many topics we should train the model to learn. Topic modelling is an unsupervised method, there is no label for the model to determine the best number of topics. Fortunately, there are a couple ways to define the optimal number of topics. Computing the topic coherence score for different numbers of topics and choosing the model that gives the highest topic coherence would be one of the ways. Perplexity and log-likelihood based V-fold cross validation are also good options for finding the optimal number of topics. However, there is no standard to say which way is the best way to pick the optimal number of topics. Sometimes, the statistical optimal number of topics may not fit the business problem well. In this paper, we use the perplexity and log-likelihood to analyze how many topics are statistically optimal.

We train LDA models with different combination of parameters with GridSearchCV and want to find the optimal number of topics and learning rate by minimizing the validation log likelihood of the model.

From the log-likelihood, it says that the optimal number of topics is around twenty to thirty. Then, we analyze the keys words under each topic to better understand each topic. Working with domain experts and leverage metadata, we can define intent domains from interpreting the keys words from each topic. The figure below shows an example plot of high frequent words under 20 topics. Using this as an example, we can see there are a couple topics about "order", "refund" and "shipping". With these information, we can define these topics belong to "Order Related" domain. There are also some other topics are about "account", "address" and "tax". These high frequent key words would give a sense of what are the documents under this topic are related to.

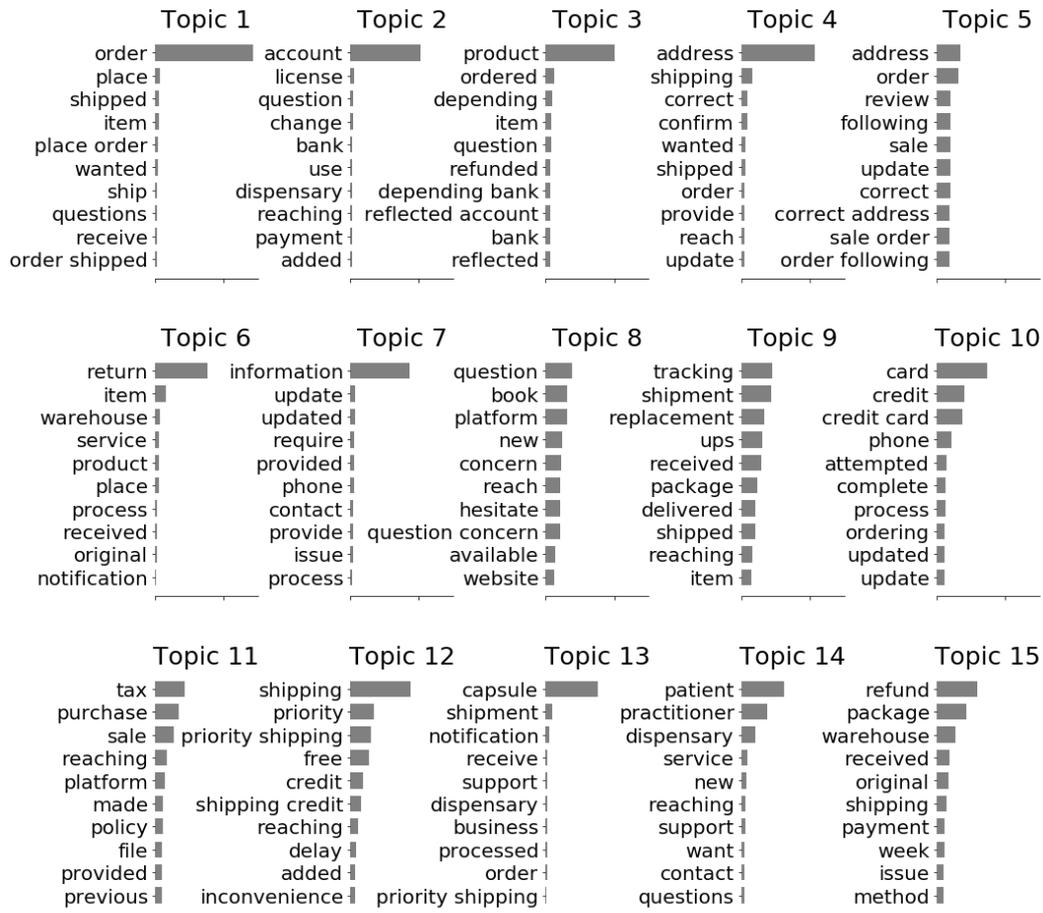

Figure 6. Keywords Under Topics

We want to emphasize the importance of domain expert involvement in defining intent domains. Topic modelling is not a perfect model that would return you a set of intent domains. But it only returns you clusters of key words. In order to give meanings to topics, we highly suggest involving domain experts since they have better understanding of meanings behind these clusters of key words.

**5.2. Intent Use Case and Its Variation Examples**

With the pre-defined intent domains, we also define a few intent use cases by analyzing the high frequent keywords of topics under each intent domain. For example, under the intent domain "Order Related", we see key words like "shipment", "return", "refund", "missing item", "address error", etc. We leverage these keywords to define intent use cases such as "Shipment issue", "return request", "refund request", "missing item', etc. There is no right or wrong to define these intent use cases. Topic modelling or other domain knowledge would be good ways to define these intent use cases.

Furthermore, for each intent use case, we also generate a couple variation examples. Since various usage of the intent use case would better capture similarity between intent use case and ticket sentences. Customer language usage usually is quite diverse. If a customer asks about closing their account, there could be various ways to ask. Therefore, generating a list of variations for each intent use case, we could measure the similarity score between these use case variations and ticket sentences to find out intent use cases that have high similarity with ticket sentences. If we use the intent use case "Close Account" as an example, we can generate a list of variations shown in the picture below.

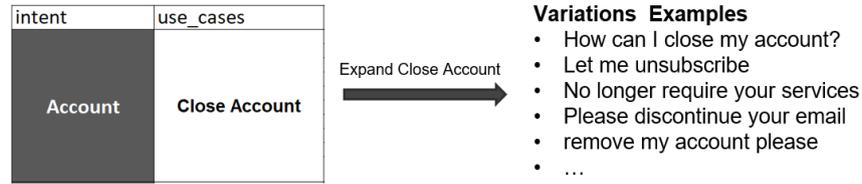

Figure 7. Example of Expanding Intent Use Case with Variations

## 6. INTENT MAPPING MODEL

In order to map the text intent with the pre-define intents use case, we initially map email/chat raw subject (if exists) to define intent domain. Then, we map the intent use case to ticket based on the cosine similarity between sentence embeddings and intent use case variation embeddings. In the case when there is no email or ticket subject, we scan through all intent domains rather than focus on single intent domain.

The reason why we leverage the subject of the ticket is because ticket subject usually has information about the user issue or intent. By leveraging ticket subject, we sometimes can narrow down the intent domain when to do intent mapping. When determining if a ticket subject is useful, ticket subject is transformed into embedding and compared similarity with intent use case embeddings. If there is a high similarity score under an intent domain, we will assign an intent domain to this ticket. Otherwise, the model will scan through all the intent domains rather than focus on one domain. Subject mapping is not necessary when there is no ticket subject or additional information about the ticket text. The goal of subject mapping is aimed to narrow down the intent domain when there exist plenty of intent domains.

When mapping the intents to ticket, we firstly split a ticket into sentences, and then compute the embeddings for each sentence and intent use case variation. During the similarity measuring process, we loop through each sentence's embeddings and each intent use case variation's embeddings, and get their cosine similarity scores. Then we will find the maximum sentence similarity scores for each intent use case, and it returns us a list of intent use cases with similarity scores. Finally, we would rank intent use case relevance based on their similarity scores. The whole intent mapping algorithm can be expressed in Algorithm 1 below.

**Algorithm 1** Document Intent Mapping
**if** $Intent - domain$ is classified **then**
    **for** L Use Cases **do**
        **for** N Sentences **do**
            Get Similarity Scores for K Variations>
    Matrix $(L, N, K)$       ▷ Similarity Score Matrix
    $(L, N) \leftarrow max(L, N, K)$   ▷ Get Max Score from K Variations
    $(L) \leftarrow max(L, N)$   ▷ Get Max Score from N Sentences
    **return** Ranked L Intents

**else if** $Intent - domain$ is not classified **then**
    **for** M Domains **do**
        **for** L Use Cases **do**
            **for** N Sentences **do**
                Get Similarity Scores for K Variations>
    Matrix $(M * L, N, K)$
    $(M * L, N) \leftarrow max(M * L, N, K)$
    $(M * L) \leftarrow max(M * L, N)$
    **return** Ranked M*L Intents

Algorithm 1. Intent Mapping Process

# 7. EVALUATION

Artwo AI also provided a limited amount of labelled data of the tickets. We are going to use these labelled data to tune our model parameters and validate the performance of our models. Both CBOW and Skip-Ngram models, and dimension 100 to 300 have been tested. Accuracy results are in the table below.

Table 2. Model Evaluation Results

| Model | Account Settings | Close Account | Account Migration | Quality Issue | Order Return | Delivery Delay | Checkout Assistance | Tax Exemption |
|---|---|---|---|---|---|---|---|---|
| Skip-Ngram (Dim=100) | **0.712** | 0.796 | **0.795** | 0.943 | 0.722 | 0.896 | **0.637** | **0.865** |
| Skip-Ngram (Dim=200) | 0.690 | **0.818** | 0.783 | **0.964** | **0.749** | **0.936** | 0.619 | 0.848 |
| Skip-Ngram (Dim=300) | 0.627 | 0.751 | 0.826 | 0.918 | 0.726 | 0.883 | 0.604 | 0.831 |
| CBOW (Dim=100) | 0.711 | 0.770 | 0.533 | 0.943 | 0.746 | 0.924 | 0.576 | 0.806 |
| CBOW (Dim=200) | 0.679 | 0.775 | 0.581 | 0.917 | 0.738 | 0.906 | 0.552 | 0.802 |
| CBOW(Dim=300) | 0.678 | 0.772 | 0.598 | 0.873 | 0.734 | 0.893 | 0.560 | 0.784 |
| Total Number of Ticket | 1620 | 378 | 3347 | 778 | 5402 | 606 | 1554 | 2587 |

# 8. CONCLUSION

From the result, we can see that even with the current pre-defined intent use case and variations, we can achieve decent accuracy on limited labelled data under a few intent domains. There is potential that the accuracy can be further improved by involving domain expert to better define the intent variations based on their experience.

By having this intent mapping model, we can quickly get an unsupervised way to understand the potential intent behind the ticket without label data. Moreover, it is easy to add additional intent domain or intent use cases by just defining a few intent variations for your intent use case. With large number of unlabelled data, we can use this word2vec model to analyze the intent behind the text data.

# ACKNOWLEDGE

We thank Alex Bakus and Alex Johns from Artwo AI for providing us data to explore and share their domain knowledge to help us better define the domain intent and intent use case.